\documentclass[onefignum, onetabnum,onealgnum]{siamonline190516}
\usepackage{comment}
\usepackage{lipsum}
\usepackage[margin=1in, footskip=36pt]{geometry}

\usepackage{endnotes}

\usepackage{titlesec}
\usepackage{titletoc}
\usepackage{pgfplotstable}
\titleclass{\task}{straight}[\section]
\newcounter{task}
\renewcommand{\thetask}{\arabic{task}}
\titleformat{\task}[hang]
    {\normalfont\LARGE\bfseries}{Task \thetask:}{1em}{}

\titleformat*{\task}{\color{header1}\bfseries}

\titlecontents{task}
              [3.8em] 
              {}
              {\contentslabel{2.3em}}
              {\hspace*{-2.3em}}
              {\titlerule*[1pc]{.}\contentspage}

\titlespacing*{\section}{0ex}{1ex}{1ex}
\titlespacing*{\subsection}{0ex}{1ex}{1ex}
\titlespacing*{\subsubsection}{0ex}{1ex}{1ex}
\titlespacing*{\paragraph}{0ex}{1ex}{1ex}
\titlespacing*{\subparagraph}{0pt}{1ex}{1ex}
\titlespacing*{\task}{0em}{1ex}{1ex}

\newcommand\blfootnote[1]{%
  \begingroup
  \renewcommand\thefootnote{}\footnote{#1}%
  \addtocounter{footnote}{-1}%
  \endgroup
}

\usepackage[sort&compress,comma,square,numbers]{natbib}

\usepackage{amsfonts,amsmath,amssymb}
\usepackage{bbm}

\usepackage{multicol}
\usepackage{enumitem}
\setlist[enumerate]{wide, labelindent=1cm,  noitemsep}
\setlist[itemize]{noitemsep}
\setlist[description]{noitemsep}

\usepackage{hhline}

\usepackage{todonotes}
\RequirePackage{ulem}

\definecolor{tableheadcolor}{gray}{0.92}
\newcommand{\topline}{ %
        \arrayrulecolor{blue1}\specialrule{0.1em}{\abovetopsep}{0pt}%
        \arrayrulecolor{tableheadcolor}\specialrule{\belowrulesep}{0pt}{0pt}%
        \arrayrulecolor{blue1}}
\newcommand{\midtopline}{ %
        \arrayrulecolor{tableheadcolor}\specialrule{\aboverulesep}{0pt}{0pt}%
        \arrayrulecolor{blue1}\specialrule{\lightrulewidth}{0pt}{0pt}%
        \arrayrulecolor{white}\specialrule{\belowrulesep}{0pt}{0pt}%
        \arrayrulecolor{blue1}}
\newcommand{\bottomline}{ %
        \arrayrulecolor{white}\specialrule{\aboverulesep}{0pt}{0pt}%
        \arrayrulecolor{blue1} %
        \specialrule{\heavyrulewidth}{0pt}{\belowbottomsep}}%

\pgfplotstableset{normal/.style ={%
        header=true,
        string type,
        font=\small,
    	column type={p{.092\textwidth}},
        every head row/.style={
            before row={\topline\rowcolor{tableheadcolor}},
            after row={\midtopline}
        },
        every odd row/.style={
            before row={\rowcolor{blue1}}
        },
        every last row/.style={
            after row=\bottomline
        },
        col sep=&,
        row sep=\midrule
    }
}

\usepackage{multirow}

\usepackage{booktabs}
\usepackage{titlesec}
\usepackage{fancyhdr}
\usepackage{fullpage}

\RequirePackage{titlesec}
\RequirePackage[font={footnotesize},{singlelinecheck=false}]{caption}
\usepackage[T1]{fontenc}
\usepackage[scaled]{helvet}
\usepackage[scaled=1.10, med]{zlmtt}

\usepackage{algorithm}
\usepackage{algpseudocode}

\providecommand{\mc}[1]{\mathcal{#1}}

\providecommand{\mb}[1]{\boldsymbol{#1}}

\providecommand{\mbb}[1]{\mathbb{#1}}

\newcommand{\jvemail}{\href{mailto:jovo@jhu.edu}{jovo@jhu.edu}}

\newcommand{\Real}{\mathbb{R}}

\DeclareMathOperator{\R}{R} 

\newcommand{\eps}{\varepsilon}

\usepackage{hyperref}       
\usepackage{url}            
\usepackage{booktabs}       
\usepackage{amsfonts}       
\usepackage{nicefrac}       
\usepackage{comment} 
\usepackage{amsmath}
\usepackage{mathtools}
\usepackage{microtype}      
\usepackage{caption, subcaption}
\pgfplotsset{compat=1.5}

\newcommand{\cT}{\mathcal{T}}
\newcommand{\cX}{\mathcal{X}}
\newcommand{\cY}{\mathcal{Y}}

\newcommand{\cH}{\mathcal{H}}
\newcommand{\cD}{\mathcal{D}}

\newcommand{\cR}{\mathcal{R}}
\newcommand{\cS}{\mathcal{S}}

\newcommand{\cF}{\mathcal{F}}

\renewcommand{\R}{\mathbb{R}}

\usepackage{todonotes}
\usepackage{setspace}
\presetkeys{todonotes}{size=\tiny}{}

\title{
Towards a theory of out-of-distribution learning
}
\author{Jayanta Dey,$^1$ 
Ali Geisa,$^{\dagger,1}$
Ronak Mehta,$^{\dagger,1}$ 
Tyler M. Tomita,$^1$
Hayden S. Helm,$^1$ 
Haoyin Xu,$^1$
Eric Eaton,$^2$ 
Jeffery Dick,$^3$  
Carey E. Priebe,$^1$ and Joshua T. Vogelstein$^{1,*}$
}
\begin{document}
\maketitle
\blfootnote{
$^{\dagger} $ denotes equal contribution,
$ ^{*} $ corresponding author: \jvemail.
}
\blfootnote{$^1$Johns Hopkins University (JHU),
$^2$University of Pennsylvania,
$^3$Loughborough University. 
}
\begin{abstract}
Learning is a process wherein a learning agent enhances its performance through exposure of experience or data. Throughout this journey, the agent may encounter diverse learning environments. For example, data may be presented to the leaner all at once, in multiple batches, or sequentially. Furthermore, the distribution of each data sample could be either identical and independent (iid) or non-iid. Additionally, there may exist computational and space constraints for the deployment of the learning algorithms. The complexity of a learning task can vary significantly, depending on the learning setup and the constraints imposed upon it. However, it is worth noting that the current literature lacks formal definitions for many of the in-distribution and out-of-distribution learning paradigms. Establishing proper and universally agreed-upon definitions for these learning setups is essential for thoroughly exploring the evolution of ideas across different learning scenarios and deriving generalized mathematical bounds for these learners. In this paper, we aim to address this issue by proposing a chronological approach to defining different learning tasks using the provably approximately correct (PAC) learning framework. We will start with in-distribution learning and progress to recently proposed lifelong or continual learning. We employ consistent terminology and notation to demonstrate how each of these learning frameworks represents a specific instance of a broader, more generalized concept of learnability. Our hope is that this work will inspire a universally agreed-upon approach to quantifying different types of learning, fostering greater understanding and progress in the field.
\end{abstract}

\tableofcontents

\section{Introduction}
\label{sec:intro}

The study of machine learning (ML) has enabled remarkable progress in artificial intelligence (AI), including revolutions in image recognition~\citep{Krizhevsky2012-sq}, natural language processing~\citep{Devlin2018-xz}, medical diagnostics~\citep{McKinney2020-ft}, autonomous control~\citep{Statt2019-ox}, and protein folding~\citep{Senior2020-ea}.
But empirical progress often outpaces theoretical understanding.  Indeed, much of the progress of the last decade in ML/AI remains to be explained~\citep{Sejnowski2020-nd}. Moreover, the vast majority of the advances in ML/AI focus on in-distribution learning, that is, learning where the training and test data are assumed to be sampled from the same distribution.

However, in the real world, such in-distribution learning problems are quite narrow in scope; we would prefer to have ML/AI solutions that can solve out-of-distribution (OOD) learning problems~\citep{Bengio2011-pt}.  
Colloquially, OOD learning operates in a regime in which training data are assumed to be sampled from a distribution that differs from the evaluation data distribution.  
As defined above, examples of OOD learning include as special cases transfer~\citep{Bozinovski1976-ij}, multitask~\citep{Bengio1992-pe,Caruana1997-mm}, meta~\citep{Baxter2000-wv}, continual~\citep{Ring1994-un}, and lifelong learning~\citep{Thrun1995-qy}. 
Current deep learning systems still struggle to adapt and generalize to seemingly trivial distributional changes~\citep{Madry2018-gy, Alcorn2019-ks}.
Moreover, as Thrun lamented in his seminal paper in 1996 introducing lifelong learning~\citep{Thrun1996-kk}, the existing learning theory was inadequate to characterize this kind of learning. 
Since then, there have been a number of efforts to formalize various OOD learning scenarios, most notably~\citet{Baxter2000-wv}, and quite recently~\citet{Arjovsky2021-mm}. However, most theoretical work on OOD learning focuses on a specific special case (such as meta or invariant learning). The lack of a unifying framework characterizing each of these different OOD learning scenarios has lead to a number of ongoing challenges. 

First, the precise goals of various algorithms are often unclear.  For in-distribution learning, the goal is always clear: minimize generalization error. Out-of-distribution learning scenarios, however, are more complex.  For example, when comparing two different OOD learning algorithms, if they both observe some in-distribution data, then one could outperform the other in terms of OOD generalization error by one of several possible mechanisms: (i) it could have better priors and/or inductive biases, (ii)  it could leverage the in-distribution data more efficiently, or (iii)  it could leverage the out-of-distribution data more efficiently.  Simply comparing accuracy fails to quantify the extent to which a given algorithm is able to leverage the OOD data; that is, it fails to quantify the magnitude of actual transfer. Therefore, measures of predictive performance, such as accuracy, are inadequate when evaluating OOD learning, if we are to understand which properties of these algorithms are doing which work.  The  evaluation criteria could instead compare performance within an algorithm when obtaining and not obtaining additional OOD data.  While many authors have proposed criteria such as forward and reverse transfer, the theoretical motivation for choosing such criteria were lacking~\citep{Ruvolo2013-hk, Lopez-Paz2017-wk, Diaz-Rodriguez2018-cz, Pearl2019-bp}. 
To give one concrete example from our own work, the efficient lifelong learning algorithm (ELLA)~\citep{Ruvolo2013-hk} specifies a clear objective function, provides convergence guarantees, and works well for shallow models with sufficiently compact task distributions (i.e., tasks that are sufficiently similar). However, ELLA does not provide any guarantees on what it will converge to, nor how it will generalize to more diverse tasks \citep{BouAmmar2015Autonomous} or deeper models \citep{Lee2019Learning}. 

Second,  the literature remains confused on a number of central issues. For example, what counts as OOD learning: must the learner perform well for any possible new task in an environment (as in meta-learning and domain generalization~\citep{Muandet2013-od}), or a specified new task (as in domain adaptation~\citep{Daume2007-cv} and covariate shift~\citep{Gretton2008-dp})? What differentiates online, continual, and lifelong learning, and how are they related to OOD learning?  Do lifelong  learners have task information at training and/or testing time~\citep{Thrun1996-kk}, or must they also infer the task itself~\citep{Finn2017-fi}? And does it count as lifelong learning if computational complexity is constant~\citep{Zenke2017-yz}, scales quasilinearly~\citep{pl}, or quadratically~\citep{kirkpatrick2017overcoming} with sample size?

We therefore revisit the foundations of learning~\citep{Shalev-Shwartz2010}, and endeavor to update them in three ways. First, we define a {generalized learning task} which includes both in-distribution and out-of-distribution (\S~\ref{sec:learning}), illustrating that in-distribution is merely a special case of out-of-distribution.  Specifically, \textbf{we  make a simple change to the classic in-distribution definition of a learning task: we no longer implicitly assume that the evaluation distribution that our learner will face at test/deployment time is the exact same distribution from which the training data are assumed to be sampled} (Figure~\ref{fig:schematic}, left).


Third, we formally define several notions of \textbf{out-of-distribution learnability}, including weak, strong, non-uniform, and consistency, to complement and extend in-distribution variants of weak, strong or probably almost correct (PAC) learnability~\citep{Valiant1984-dx},  non-uniform learnability~\citep{Shalev-Shwartz2014-zo}, and consistency~\citep{bickel-doksum} (\S~\ref{sec:gen_learnability}).
We then prove the relationship between each of these generalized notions of learnability. 
There are three conceptual differences between our definition of learning and previous definitions. First, in previous definitions, the evaluation distribution and training data distribution were (often implicitly) assumed to be the same. Here, those two distributions need not be related (though they must be related to solve the problem).  Second, {our explicit goal is to \textit{improve} performance with (data)}, rather than, say, achieve (Bayes) optimal performance---a much more modest aspiration. Third, we do not require large sample sizes. Rather, our definition includes zero-shot and few-shot learning.  Note that these three relaxations of existing formal definitions of learning  improve the alignment between machine and natural (biological) learning, which partially motivated this work.

\begin{figure}[t!]
    \centering
    \includegraphics[width=\linewidth]{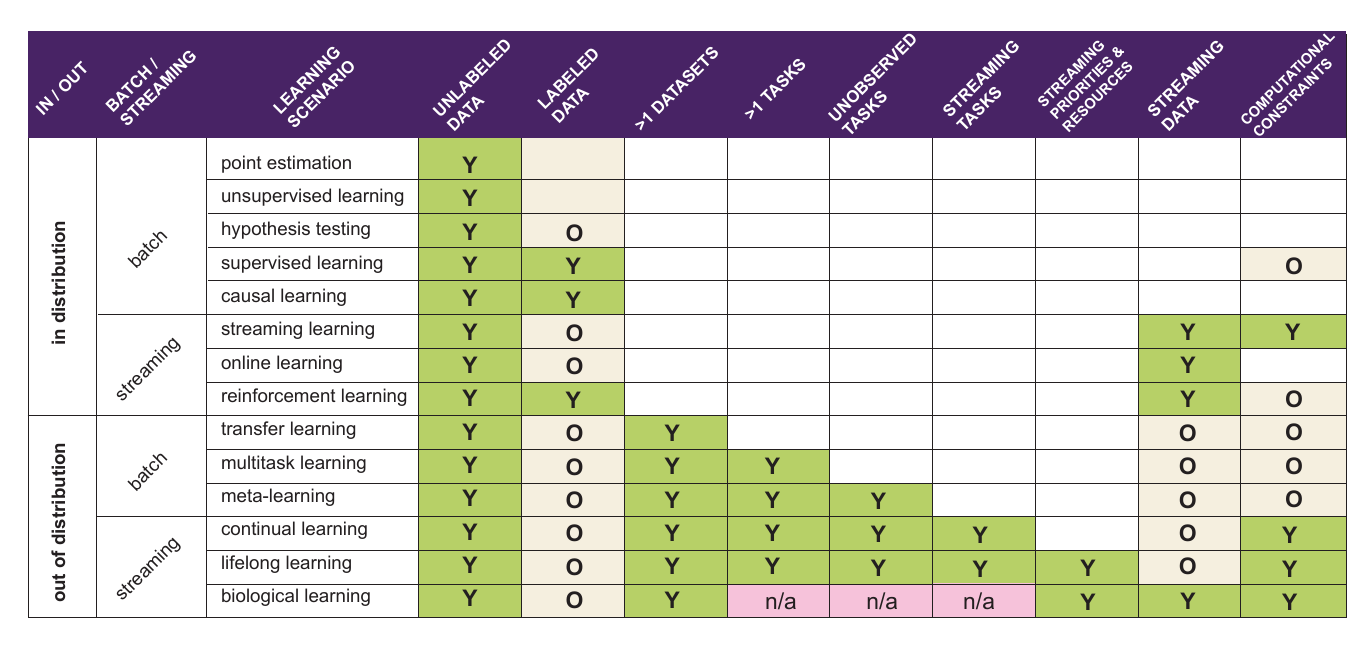}
    \caption{The characteristics of different learning paradigms. Y: present, O: optional, n/a: not applicable.
    }
    \label{fig:schematic_table}
\end{figure}

Fourth, we leverage our definition of learning, learnability, and learning efficiency to formalize, quantify, and hierarchically organize learning in multiple distinct OOD learning scenarios, including transfer, multitask, meta, continual, and lifelong learning (\S~\ref{sec:hier} and Figure~\ref{fig:schematic}, right). Doing so required formalizing some conventions that are not yet universally agreed upon. While some readers may disagree with some of the choices we made in defining this hierarchy, our main point is not the particular hierarchy \textit{per se}.
Rather, our primary intention is to illustrate the flexibility of our formalism; specifically, that it provides a single coherent lens through which essentially any learning scenario (including both in-distribution and out-of-distribution learning) may be evaluated, much like generalization error provides for in-distribution learning (Figure~\ref{fig:schematic_table}).

Finally, Thrun  pointed out in 1996 that lifelong learning  is essential for human learning~\citep{Thrun1996-kk}.  
Moreover, it seems essential for biological learning more generally, not just human learning~\citep{Marcus2019-dj}.  For example, honey bees with only $\approx$1 million neurons can learn `same versus different' tasks, and moreover, they can perform zero-shot cross-model transfer learning~\citep{Giurfa2001-nr}. Other species with larger brains can perform all sorts of tasks that remain outside the realm of modern ML/AI~\citep{Lake2017-ip}. We believe that OOD learning is the key capability that biological learning agents leverage to achieve natural intelligences that surpass modern ML/AI~\citep{Goyal2020-qs}. And therefore, we hope our contribution will help bridge the gap between biological and machine learning and intelligence,  
which was one of Valiant's originally stated goals, as implied by the first sentence of his seminal paper:
    ``Humans appear to be able to learn new concepts without needing to be programmed explicitly in any conventional sense.''~\citep{Valiant1984-dx}


\section{A unified framework for  learning}
\label{sec:prelims}

\label{sec:learning}

Here we formally define learnability, starting with in-distribution learnability,  leveraging the framework proposed by \citet{Glivenko1933-wm} and~\citet{Cantelli1933-vj} nearly 100 years ago (and was  further refined by~\citet{Vapnik1971-um} and~\cite{Valiant1984-dx} 50 years ago). We then illustrate how it can be generalized to include many variants of out-of-distribution learnability, including transfer, multitask, meta, continual/lifelong, and prospective learning. 
Our key contribution is providing a coherent terminology and notation that illustrates that each of these learning frameworks are special cases of a more general notion of learnability.
We follow the notation established by~\citet{Shalev-Shwartz2010} and~\citet{Fokoue2020-sv} for in-distribution learning, departing as appropriate to generalize and include various forms of out-of-distribution learning. 
We introduce a sequence of increasingly more complicated learning scenarios. Each scenario builds on the previous scenarios. 
We use lowercase, capital and calligraphic letters to denote samples, random variables and sets, respectively. For example, $a$ is a sample of the random variable $A$, and $\mc{A}$ is the set of possible samples $a$. 


\subsection{Decision Task}
\label{sec:task}

In this section we introduce the decision task (adapted from \citet{bickel-doksum}, Chapter $1.3$), which contains several important notions---like Bayes hypothesis and Bayes risk---which we will use later in our framework. The decision problem is a mathematical formulation of an ideal goal that a learner would try to achieve.

\begin{itemize}

    \item \textbf{Data Space} $\cS = \mc{X} \times \mc{Y}$ is the product space of the set of inputs, questions or queries, $\mc{X}$, and the set of outputs, actions, responses, or predictions, $\mc{Y}$.

    
    
   \item \textbf{Hypotheses} $\cH \subseteq \{h \mid h: \cX \to \cY \}$ is a (potentially strict) subset of all mappings from the \textit{input space} $\mc{X}$ to the \textit{output space} $\mc{Y}$.~\footnote{Because $\cH$ is not necessarily the set of all possible mappings from $\mc{X}$ to $\mc{Y}$, we will sometimes refer to $\cH$ as the set of feasible hypotheses. The constraints on the considered hypothesis result from the nature of the problem and desired solutions.}
   
    \item  \textbf{Distributions} $\cD$. We assume that the  data $(x,y)$ are realizations of $(\cX,\cY)$-valued random variables $(X,Y)$, which are drawn according to some true but unknown distribution $D \in \cD$. 
    
    \item \textbf{Risk} $R : \mc{H} \times \cD \mapsto \R_{\geq 0}$. A risk evaluates the performance of a hypothesis with respect to a distribution. 
    Often, implicit in the risk functional, is a loss function $\ell$. For example, for supervised learning $\ell: \mc{Y} \times \mc{Y} \mapsto \mbb{R}_{\geq 0}$, and the risk is defined as the expected loss with respect to $D$:     $R(h) = \mbb{E}_{(X, Y) \sim D}[\ell(h(X), Y)]$.
\end{itemize}

     Given these components, we can define a {\bf decision task} (which has no data, and therefore, nothing to learn):

\begin{definition}[Decision Task]
    A \textit{decision task} has the form 
    \begin{equation} \label{eq:problem}
    \begin{array}{ll}         
         \textnormal{minimize}  &R(h) 
        \\ \textnormal{subject to }\,  &h \in \cH
    \end{array} \enspace .
    \end{equation}
\end{definition}

Let $h^*$ denote a \textbf{Bayes optimal hypothesis}, that is, a hypothesis that solves Objective \eqref{eq:problem} ignoring constraints on $\cH$.
Let $R^* = R(h^*)$ denote the \textbf{Bayes optimal risk}.
Let $h^\diamond$ denote the optimal solution to Objective \eqref{eq:problem}, respecting any constraint on $\cH$. Let $R^\diamond = R(h^\diamond)$. 
Note that neither $h^*$ nor $h^\diamond$ needs to be unique, and that neither $h^\diamond$ nor $R^\diamond$  needs to exist (in which case we could be looking for an infimum rather than a minimum). The \textbf{excess risk} or \textbf{model error} is defined by $R^\diamond-R^*$~\cite{Vapnik1998-po}.

\subsection{PAC In-Distribution Learning}

In \textbf{decision task}, we are given a hypothesis set $\mc{H}$ from which we aim to choose the optimal $h$. In \textbf{learning tasks}, a {\bf learner} uses $n$ random data points to search for an $h$ that minimizes risk, $R(h)$. Defining a learning task formally requires a few additional concepts.
\begin{itemize}

    \item \textbf{Data} $s_n \in \cS$ where $\cS = \{s_n \mid  n \in \mbb{Z}_{\geq 0}\}$, where $s_n$ is a set of $n$ data points. For example, in supervised learning problems, $s_n$ is a set (or multiset) of $n$ pairs in $\cX \times \cY$, for example: $s_n = ( (x_1,y_1), \cdots, (x_n,y_n))$. However, more generally, the elements in $s_n$ need not be input-output pairs; for example, in unsupervised learning samples might be purely in $\cX$.  Note that for $n=0$, $s_0$ refers to the empty set. 

   \item \textbf{Hypotheses} $\cH$ is defined as above.
   
    \item \textbf{Distributions} $\mc{D}=\{\cD_n | n \in \mbb{Z}_{\geq 0}\}$, is the set of all possible distributions of the data, that is, $S_n \sim D_n \in \cD_n$, where $\cD_n$ is the set of all possible distributions generating $n$ samples. 

    \item \textbf{Risk} $R$ is defined as above.

    \item \textbf{Learners} $\cF \subseteq \{f: \cS \times \Lambda \mapsto \cH\}$,  where $f$ is a learner, and $\cH$ is the set of hypotheses that any $f \in \cF$ can yield.\footnote{
    In general, the $\cH$ that learners map to and the $\cH$ defining the set of feasible hypotheses (defined in the decision problem) need not be the same.  For example, the decision problem may be looking for only linear decision rules, but in learning problem, we could use a Quadratic Discriminant learner, which may yield a nonlinear as well as a linear decision rule.  Thus, in general, for a learning problem to be well specified, these two $\cH$'s must be overlapping.
    If the feasible $\cH$ contains $h^*$, with enough training samples and certain assumptions on $\cD$ being satisfied we can achieve Bayes optimal solution to Objective \eqref{eq:problem}. Throughout, we assume the two $\cH$'s are the same.
}
The first input to the learner is the training data, $s_n$.%
The second input is the hyperparameter, $\lambda \in \Lambda$, which can  incorporate prior knowledge, initial conditions, and potentially other kinds of side information~\cite{Shannon1958-no}. 
We will typically suppress the $\Lambda$-valued inputs for notational brevity, so we write $f(s_n) = \hat{h}_n$, and assume that any two different hypotheses, $h$ and $h'$ share the same hyperparameters unless otherwise stated.  


\end{itemize}

To formalize a learning task, we introduce a few constants.  A precision parameter, $\eps$, controls how much we require to learn.  If we have a larger $\eps$, that means we are requiring more learning to constitute `learning'.  The upper bound on how much one can learn is given by the difference between `chance performance', $R(h_0)$, and optimal performance, $R^\diamond$.  The reason we call $R(h_0)$ chance performance is because it corresponds to the lowest risk one would obtain given no data whatsoever.  Therefore, $0 < \eps < R(h_0) - R^\diamond $.  A confidence parameter, $\delta$, controls how confident we must be about our performance.  For historical reasons, we require that the probability we learn by at least $\eps$ is bounded below by $1-\delta$.  Thus, smaller $\delta$ implies that we require more confidence. Therefore, $0 < \delta < 1$.  Usually, we will be interested in cases where $\delta < 0.5$, corresponding to cases where it is more likely that we learn. Given these parameters, we can now define an in-distribution learning task.

\begin{definition}[In-Distribution Learnability] \label{def:ID_learn}
We say that a learner $f$ $(\eps, \delta, n)$-learns from $D_n$ relative to $\mc{H}$ if, when given $n$ iid samples from $D_n$, with probability at least $1-\delta$, $f$ outputs a hypothesis $\hat{h}_n$ that performs at least $\eps$ better than the best hypothesis learned using no data, $h_0$:
\begin{equation}\label{eq:id_learn}
    \mbb{P}[R(h_0) - R(\hat{h}_n) \geq \eps] \geq 1 - \delta.
\end{equation}
\end{definition}

An in-distribution learning task is to find an $f \in \cF$ that satisfies the conditions in~\ref{def:ID_learn}.
Some remarks about the above:
\begin{itemize}
    
    \item PAC learning is typically defined relative to an optimal hypothesis, rather than chance, and for all choices of $\eps$ and $\delta$, rather than some. In PAC learning, defining learning relative to chance or optimal is equivalent,  because of the weak learner theorem~\cite{Kearns1994-yc}.  We therefore use chance here to be consistent with the more general notions of learning defined below.

    \item Note that $\mbb{P}$ in \eqref{eq:id_learn} denotes joint probability over the random draw of $n$ data points.  Implicit in the above is that we used $n$ sample points to learn $\hat{h}_n$. To be `in-distribution' requires that we make certain assumptions about the relationship between the distribution of the particular data points we sampled, $s_n$, and all other potential data, $\mc{S} \setminus \{s_n \}$. Typically, the assumption is that all the data are mutually independent and sampled from identical distributions ($iid$), that is, $D_n = \prod_n D$ for all $n \in \mbb{Z}_{\geq 0}$.  Therefore, the risk above is with respect to the same distribution, $D$.

\end{itemize}

\subsection{PAC Transfer Learning}

In transfer learning, we have two distinct kinds of data: source data and target data.  Below, we describe what this means, and how to generalize PAC in-distribution learning to this scenario. 

\begin{itemize}

    \item  \textbf{Data} $\cS$  is defined as above, except that each data point is augmented with a label $z_i$ to indicate whether it is source or target data, corresponding to $z_0$ or $z_1$, respectively. 
    For example, if the data are supervised, then  $s_n = ( (x_1,y_1, z_1), \cdots, (x_n,y_n,z_n))$.
    The source ($\mc{X}^0 \times \mc{Y}^0$)  and the target ($\mc{X}^1 \times \mc{Y}^1$) data spaces may differ. 
    Therefore, we define $\mc{X} = \mc{X}^0 \cup \mc{X}^1$, and $\mc{Y} = \mc{Y}^0 \cup \mc{Y}^1$. Let $n_0$ and $n_1$ denote the source and target sample sizes, respectively, so $n=n_0+n_1$, and $\mb{n}=[n_0,n_1]$. 
        
    \item \textbf{Hypotheses} $\mc{H} = \{ h \mid h : \mc{X}^1 \mapsto \mc{Y}^1\}$ is the set of hypotheses that map from the target input space to the target output space. 
    
    \item  \textbf{Distributions} $\mc{D}$ is defined as above, except that each $D_n$ must be the joint distribution of $n$ samples including  the random variable $Z$ indicating whether a data point is in the source or target distribution. An implication of this is that each $D_n$ can be factorized into a conditional distribution (conditioned on whether the data is source or target) and a prior (specifying the probability of sampling a source or target point).  


    \item \textbf{Risk} $R : \mc{H} \times \mc{D} \mapsto \Real_{\geq 0}$ is defined as above, except it is only with respect to the target distribution. 

    \item \textbf{Learners} $\mc{F}$ are technically the same as those for in-distribution learning, with the caveat that the data space $\mc{S}$ and the hypotheses $\mc{H}$ are defined based on the above generalized definitions.  Moreover, learners in transfer learning have a different role: their job is to use all the \textit{source} and the \textit{target} data to improve performance on the \textit{target} distribution. With a slight abuse of notation, we let $\hat{h}_{n_1}$ denote the hypothesis learned only on the $n_1$ target samples. 
\end{itemize}

\begin{definition}[Transfer Learnability] \label{def:transfer_task}
We say that a learner $f$ $(\eps, \delta, \mb{n})$-learns from $D_n$ relative to $\cH$ if, when given $n$ samples from $D_n$ (with $n_1 \geq 0$ samples from the target distribution), with probability at least $1-\delta$, $f$ outputs a hypothesis $\hat{h}_n$ that performs at least $\eps$ better than the hypothesis learned using only the target data, $\hat{h}_{n_1}$:
\begin{equation}\label{eq:trx_task}
    \mbb{P}[R(\hat{h}_{n_1}) - R(\hat{h}_n) \geq \eps] \geq 1 - \delta.
\end{equation}
\end{definition}

A transfer learning task is to find an $f \in \cF$ that satisfies the conditions in~\ref{def:transfer_task}.
Some remarks about the above:
\begin{itemize}
    \item The outer probability $\mbb{P}$ is evaluated on random draws of $n$ samples including $n_1$ target and $n_0$ source samples.

    \item When all the data are mutually $iid$---meaning source and target distributions are the same, and all data are independent---and 
    $n_1=0$, transfer learning reduces to in-distribution learning.  Thus, transfer learning is an explicit generalization of in-distribution learning. 

    \item In transfer learning, we only need to define a single risk. In other words, the source data might not be associated with a task at all, that is, the risk is not evaluated on the source distribution.  
\end{itemize}

\subsection{PAC Multitask Learning}

In multitask learning problems, there may be multiple data sets with an environment of  multiple tasks, $\cT = \{1, 2, \cdots, T\}$.  For this reason, we first augment the aforementioned components of a learning task,  to be able to account for multiple tasks.

\begin{itemize}
    \item \textbf{Data} $\mc{S}$ is defined as above, except each $z_i$ is no longer binary, rather, it is a categorical indicator that the data point is an element of one of $T$ data sets where each dataset is specifically associated with the target distribution of a particular task (and typically $T<\infty$).\footnote{More generally, we may have $J$ datasets, where $J \neq T$ and each dataset may be associated with the target distributions of multiple tasks. For simplicity, we do not consider such scenarios further at this time.}  Let $n_t$ denote the number of data points in dataset $t$, so that $n=\sum_{t=1}^T n_t$ and $\mb{n}=[n_1,n_2,\ldots, n_T]$.

    \item \textbf{Hypotheses} $\mc{H} \subseteq  \{ h = \bigcup_{t \in \mc{T}}  h^t \mid h^t : \mc{X}^t \mapsto \mc{Y}^t \,\, \forall t \in \mc{T} \}$ is the set of hypotheses, where each hypothesis is a union of hypotheses from all the tasks. 

    \item \textbf{Distributions} $\mc{D}$ is defined as above, except that each $D_n$ must be a joint distribution including the categorical indicator for which of the $T$ datasets a data point is sampled from. As above, each $D_n$ can be factorized into a product of a conditional distribution (conditioned on a dataset) and a prior (specifying the likelihood of sampling from that dataset). 
    The input and output spaces for each data set could be different or not, as in transfer learning.

    \item \textbf{Risk}  $\cR = \{R_1, \ldots, R_T\}$ is the set of risks, one for each task $t \in \mc{T}$.

    \item \textbf{Learner} $\mc{F}$ are again technically the same as the above, with the caveat that the data and hypothesis spaces are both significantly more complex. Continuing our slight abuse of notation, we let $\hat{h}_{n_t}$ denote the hypothesis learned only on the $n_t$ samples from Task $t$.  

\end{itemize}

\begin{definition}[Multitask Learnability] \label{def:multi_task}
We say that a learner $f$ $(\eps_t, \delta_t, \mb{n})$-learns from $D_n$ relative to $\cH$, if for each $t \in \cT$, when given $n$ samples from $D_n$, with probability at least $1-\delta_t$, $f$ outputs a hypothesis $\hat{h}_n$ that performs at least $\eps_t$ better than the hypothesis learned using only the target data associated with the $t^{th}$ task, $\hat{h}_{n_t}$: 
\begin{equation}\label{eq:trx_multi}
    \mbb{P}[R_t(\hat{h}_{n_t}) - R_t(\hat{h}_n) \geq \eps_t] \geq 1 - \delta_t.
\end{equation}
A multitask learning task is to find an $f \in \cF$ that satisfies the conditions in~\ref{def:multi_task}.
\noindent Multitask learning is therefore transfer learning between $T$ different tasks.  If we happen to only care about a single task $t$, we would set $\mc{T}=\{t\}$ in Definition~\ref{def:multi_task},
and then multitask learning  reduces to transfer learning.  Therefore, multitask learning is an explicit generalization of transfer learning.  

\end{definition}
%


\subsection{PAC Meta-Learning}
Loosely, meta-learning is learning to learn~\citep{Thrun1995-qy}. The learner has some past experiences, and learns to learn only if these past experiences help it learn in new situations more effectively.  The data, hypotheses, distributions, risk, and learners are defined as in multitask learning. 







\begin{definition}[Zero-Shot-Learnability] \label{def:zero_shot_learn}
    Given $t-1$ tasks, we say that a learner $f$ $(\eps, \delta, \mb{n})$-learns for the $t^{th}$ task from $D_n$ relative to $\cH$ if, when given $n$ samples from $D_n$, with probability at least $1-\delta$, $f$ outputs a hypothesis $\hat{h}_n$ that performs at least $\eps$ better than the best hypothesis using no data, $h_0$:
    \begin{equation}\label{eq:zero_shot_learn}
        \mbb{P}[R_{t}(h_0) - R_{t}(\hat{h}_n) \geq \eps] \geq 1 - \delta.
    \end{equation}
\end{definition}
A zero-shot-learning task is to find an $f \in \cF$ that satisfies the conditions in~\ref{def:zero_shot_learn}.

\begin{definition}[Few-Shot-Learnability] \label{def:k_shot_learn}
    Given $t$ tasks, we say that a learner $f$ $(\eps, \delta, \mb{n})$-learns for the $t^{th}$ task from $D_n$ relative to $\cH$ if, when given $n$ samples from $D_n$, with probability at least $1-\delta$, $f$ outputs a hypothesis $\hat{h}_n$ that performs at least $\eps$ better than the hypothesis learned using only $t^{th}$ task data, $\hat{h}_{n_t}$:
    \begin{equation}\label{eq:k_shot_learn}
        \mbb{P}[R_{t}(\hat{h}_{n_t}) - R_{t}(\hat{h}_n) \geq \eps] \geq 1 - \delta.
    \end{equation}
\end{definition}
A few-shot-learning task is to find an $f \in \cF$ that satisfies the conditions in~\ref{def:k_shot_learn}. 
\noindent

Zero-shot learning is a special case of few-shot learning, when $n_t=0$.  Note that few-shot learning is the same as transfer learning, though typical applications of few-shot learning concern the case when $n_t \ll n$, and often $n_t < 10$. Because of these equivalences, few-shot learning is also a special case of multitask learning.

\subsection{PAC Streaming Learning}
Streaming learning is similar to in-distribution learning introduced before. The difference is that data points arrive sequentially, so we no longer assume we have the whole dataset in the beginning. To ensure that a streaming learner does not merely store all past data and retrain from scratch, we introduce computational constraints using the standard `little-o' notation. Formally, $f(n) \in o(g(n))$ implies that for any constant $\eta>0$ there exists $n'$ such that $|f(n)| \leq \eta g(n)$ for all $n>n'$.  Colloquially, little-o means ``grows strictly slower than''. 

\begin{itemize}

    \item  \textbf{Data} $s_n \in \cS$,  $s_n$ is now a sequence, rather than a set or multiset. ~\footnote{At a time point, the learner may receive a batch of data samples rather than a single sample. The sequence of samples does not matter within a batch. For simplicity, we consider each sample arrives sequentially.} 
        
    \item \textbf{Hypotheses} Same as in-distribution learning. Note that in streaming learning, one can use hypothesis to save the past data samples. Hence, typically in streaming learning, we impose space constraints on the hypotheses, for example, $h \in O(1)$.
    
    \item  \textbf{Distributions} $\cD=\{ \mc{D}_n \mid n \in \mathbb{Z}_{\geq 0} \}$, but now each $D_n$ is a distribution of sequences (rather than sets) of length $n$.  For \textit{fixed} streaming learning, we assume that each $D_n = \prod_n D$, that is, the samples are still iid.  For dynamic streaming learning, we still assume the samples are independent, but we assume that at each time step, the data might be sampled from a different distribution. For more elaborate streaming learning settings, we may assume the data are sampled from a Markov process or some more general stochastic process.

    \item \textbf{Risk} $R : \mc{H} \times \mc{D} \mapsto \Real_{\geq 0}$ is defined as above in the in-distribution learning.

    \item \textbf{Learners} $\mc{F} \subseteq \{f: \cS \times \cH \times \Lambda \mapsto \cH \mid f_n \in o(n^2)\}$ is different from the in-distribution learning in a few ways.  First, at each time step, it only gets a sample.  
    Second,  the learner can use the past hypothesis $\hat{h}_{n-1}$ to update it to estimate $\hat{h}_n$ using the current sample. Third, there are computational constraints on the learner as well, for example, $f \in o(n^2)$.
\end{itemize}

\begin{definition}[Streaming Learnability] \label{def:streaming_learn}
    We say that a streaming learner $f$ $(\eps, \delta, n)$-learns from $D_n$ relative to $\mc{H}$ if, when sequentially given $n$ iid samples from $D_n$, with probability at least $1-\delta$, $f$ outputs a hypothesis $\hat{h}_n$ that performs at least $\eps$ better than the best hypothesis using no data, $h_0$:
    \begin{equation}\label{eq:streaming_learn}
        \mbb{P}[R(h_0) - R(\hat{h}_{n}) \geq \eps] \geq 1 - \delta.
    \end{equation}
\end{definition}
A streaming learning task is to find an $f \in \cF$ that satisfies the conditions in~\ref{def:streaming_learn}.

\subsection{PAC Continual/Lifelong Learning}
Loosely, continual learning is sequential learning to learn, with the caveat that we also desire to not forget (or improve performance on past tasks).

\begin{itemize}
    \item \textbf{Data} $\cS$ is same as defined above in the streaming learning, except that the data across tasks is sampled sequentially, and the data within a task may also be sampled sequentially. Also, the number of total tasks so far, $T_n$ may be unbounded, but necessarily cannot grow faster than sample size $n$.  We define $T$ as the limit of the total number of tasks, and so, $\lim_{n \to \infty} T_n \to T$, where $T$ may or may not be finite. The data may (task aware) or may not (task unaware) contain the categorical indicator for task identity. 

    \item \textbf{Hypotheses} For each new $n^{th}$ sample, we have 
        $\mc{H}_n \subseteq  \{ h = \bigcup_{t \in \mc{T}}  h^t \mid h^t : \mc{X}^t \mapsto \mc{Y}^t \text{ and } h \in o(n^2) \}$, and we have $\mc{H}_{n-1} \subseteq \mc{H}_n$. 

    \item \textbf{Distributions} $\cD=\{ \mc{D}_n \mid n \in \mathbb{Z}_{\geq 0} \}$ same as multitask learning, but now each $D_n$ is a distribution of sequences.

    \item \textbf{Risk}  is defined as in multitask learning above.

    \item \textbf{Learner} $\cF = \{\cF_n| n \in \mbb{Z}_{\geq 0}\}$ With new samples, the learner uses the current data and past hypothesis to obtain a new learner $\cF_n \subseteq \{f_n: \cS \times \mc{H}_{n-1} \times \Lambda \mapsto \cH_n \mid f_n \in o(n^2)\}$. The  differences from the above are that $f_n$ also uses past hypotheses, and has computational constraints. 

\end{itemize}

Note that continual learning is neither a generalization or a specialization of multitask learning. This is because it considers a reduced space of hypotheses and learners compared to that of multitask learners (due to the computational constraints), and also have a gradually expanding space of hypotheses (due to the sequential nature of the learners which update prior hypotheses, rather than learning them \textit{de novo}). Conceptually, continual learning lies somewhat between streaming and multitask learning. Because learning is sequential, besides having one overall notion of learning, we have two, namely forward (learning from the past) and backward (learning from the future) learning:

\begin{definition}[Forward Learnability] \label{def:forward_learn}
After having observed a total of $t$ tasks, we say that a continual learner $f$ $(\eps, \delta, n)$-forward-learns for the $t^{th}$ task from $D_n$ relative to $\mc{H}$ if, when sequentially given $n$ samples from $D_n$, with probability at least $1-\delta$, $f$ outputs a hypothesis $\hat{h}_n$ that performs at least $\eps$ better than the hypothesis learned using only the target data associated with the $t^{th}$ task, $\hat{h}_{n_t}$:
    \begin{equation}\label{eq:forward_learn}
        \mbb{P}[R_{t}(\hat{h}_{n_t}) - R_{t}(\hat{h}_{n}) \geq \eps_t] \geq 1 - \delta_t.
    \end{equation}
\end{definition}
 Continuing our slight abuse of notation, we let $\hat{h}_{n_t}$ denote the hypothesis learned only on the $n_t$ samples from task $t$.
 

\begin{definition}[Backward Learnability] \label{def:backward_learn}
    After having observed a total of $t$ tasks, we say that a continual learner $f$ $(\eps_{t'}, \delta_{t'}, n)$-backward-learns for any task $t' < t$ from $D_n$ relative to $\mc{H}$ if, when sequentially given $n$ samples from $D_n$, with probability at least $1-\delta_{t'}$, $f$ outputs a hypothesis $\hat{h}_n$ that performs at least $\eps_{t'}$ better than the hypothesis learned using the data associated with and up to the $t^{th}$ task, $\hat{h}_{n'}$:
        \begin{equation}\label{eq:backward_learn}
            \mbb{P}[R_{t'}(\hat{h}_{n'}) - R_{t'}(\hat{h}_{n}) \geq \eps_{t'}] \geq 1 - \delta_{t'}.
        \end{equation}
         With a slight abuse of notation, we let $h_{n'}$ denote the hypothesis learned using all the data up to and including the sample size when the learner detects Task $t'$, i.e., $n' = \sum_{i=1}^{t'} n_i$.
    
\end{definition}

Combining the two types of learning defined above, we define an overall notion of continual learning:

\begin{definition}[Overall/continual Learnability] 
\label{def:overall_learn}
After having observed a total of $t$ tasks, we say that a continual learner $f$ $(\eps_{t'}, \delta_{t'}, n)$-learns for any task $t' \leq t$ from $D_n$ relative to $\mc{H}$ if, when sequentially given $n$ samples from $D_n$, with probability at least $1-\delta_{t'}$, $f$ outputs a hypothesis $\hat{h}_n$ that performs at least $\eps_{t'}$ better than the hypothesis learned using only the target data associated with the $t'$-th task, $\hat{h}_{n_{t'}}$:
        \begin{equation}\label{eq:overall_learn}
            \mbb{P}[R_{t'}(\hat{h}_{n_{t'}}) - R_{t'}(\hat{h}_{n}) \geq \eps_{t'}] \geq 1 - \delta_{t'},
        \end{equation}
\end{definition}
A continual learning task is to find an $f \in \cF$ that satisfies the conditions in~\ref{def:overall_learn}.

\section{Generalized definitions of learnability}
\label{sec:gen_learnability}

A learning task defines the goal of a learner.  Given such a definition, it is natural to wonder whether a particular learner achieves that goal, i.e., whether a task is \textit{learnable} for a learner.  More generally, we may desire to know the extent to which a learner achieves that goal.  In classical in-distribution learning theory, there are many formal, complementary definitions of learning, including weak learnability~\citep{Kearns1994-yc}, strong learnability~\citep{Schapire1990-zz}, non-uniform learnability~\citep{Shalev-Shwartz2010}, and (universal) consistency~\citep{bickel-doksum}.
None of these definitions of learning, however, are sufficiently general to account for all the modern learning paradigms we study in ML/AI, including transfer, multitask, meta, continual, and lifelong learning.  For that reason, here we extend those definitions of learning to include the various out-of-distribution learning scenarios mentioned above.  




\subsection{Generalized Learnability}

There are several well established definitions of learnability available in the in-distribution literature, including 
including weak~\citep{Kearns1994-yc}, strong~\citep{Schapire1990-zz}, and non-uniform learnability~\citep{Shalev-Shwartz2010}, as well as (universal) consistency~\citep{bickel-doksum}.  In this subsection, we illustrate how to generalize these notions of learnability under generalized learning problem. 


\begin{definition}[Weak Learnability] \label{def:weak-ood}
A hypothesis class $\cH$ is weakly learnable with respect to a target distribution $D$ with precision $\epsilon$ if there exists a learner $f \in \cF$ such that for any confidence $\delta$, there exists $M(\frac{1}{\epsilon}, \frac{1}{\delta}) \in \mbb{N}$ and for $n \geq M$, $f$ outputs a hypothesis $\hat{h}_n \in \cH$ such that:

\begin{equation}
    \mbb{P}[R(\hat{h}_0) - R(\hat{h}_{n}) \geq \epsilon] \geq 1 - \delta,
\end{equation}
\end{definition}


\noindent 
where $\hat{h}_0 = f(s_0)$ and $s_0$ is an empty data set.

\begin{definition}[Strong Learnability]
A hypothesis class $\cH$ with precision $\epsilon$ is strongly learnable with respect to a target distribution $D$ if there exists a learner $f \in \cF$ such that for any confidence $\delta$, there exists $M(\frac{1}{\epsilon}, \frac{1}{\delta}) \in \mbb{N}$ and for $n \geq M$, $f$ outputs a hypothesis $\hat{h}_n \in \cH$ such that:
\begin{equation}
    \mbb{P}[R(\hat{h}_n) - R^\diamond \leq \eps] \geq 1 - \delta,
\end{equation}
\end{definition}

\begin{figure}[b!]
    \centering
    \includegraphics[width=\linewidth]{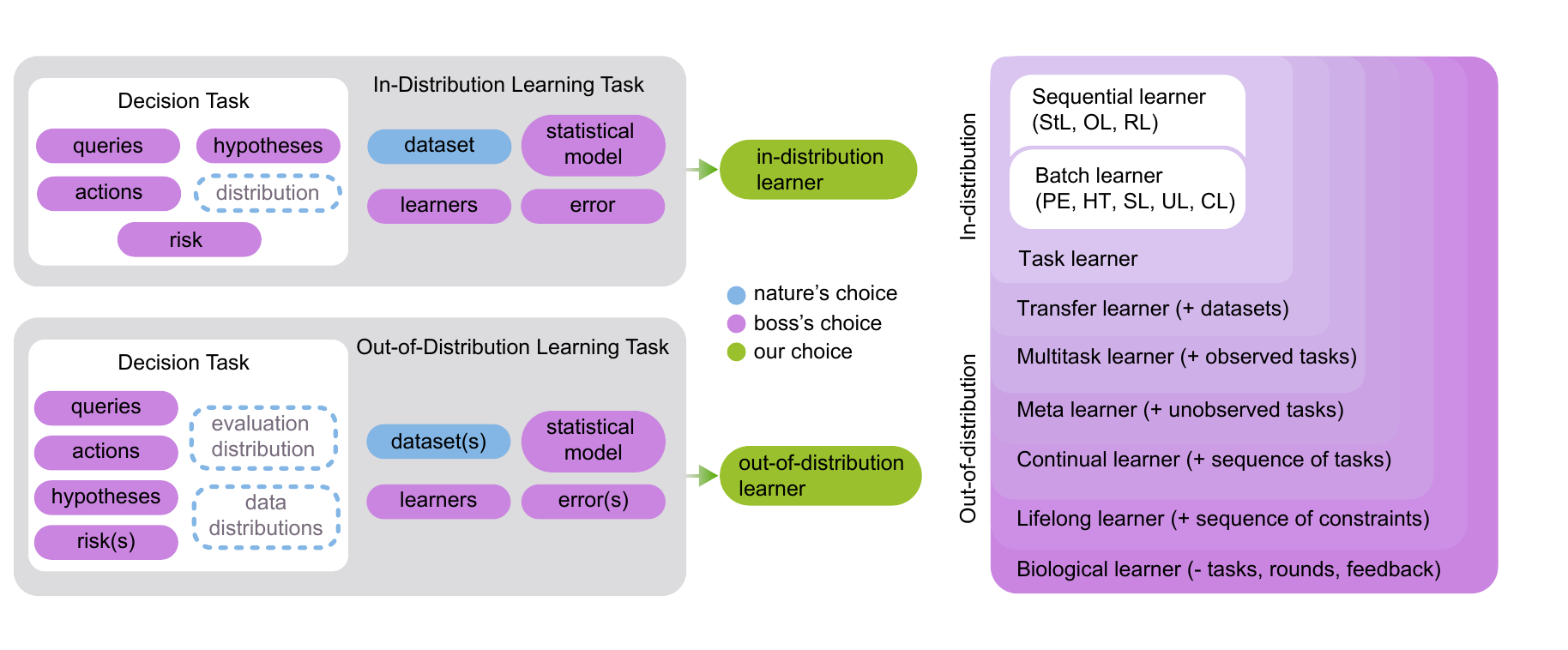}
    \caption{(Left) Decision tasks (top) are composed of five components, and the goal is to choose a hypothesis based on the known distribution that minimizes risk.  In an in-distribution learning task (middle), the distribution is not available, so a feasible in-distribution learner must leverage a data set, to find a hypothesis that minimizes error under an assumed statistical model.  In out-of-distribution learning tasks (bottom), the distribution over queries need not be about the assumed distribution of the data, and there may be multiple data sets, risks, and errors. Each component is provided by one of three different actors: nature, the boss, or the machine learning practitioner.
    (Right) Schematic illustrating the nested nature of learning problems. PE = point estimation; HT = hypothesis testing; SL = supervised learning; UL = unsupervised learning; CL = causal learning; StL = streaming learning; OL = online learning; RL = reinforcement learning.
    }
    \label{fig:schematic}
\end{figure}

\section{Discussion}
\label{sec:disc}

\subsection{A high-dimensional landscape of kinds of learning problems}

A number of previous papers have endeavored to developed taxonomies of transfer and/or continual learning~\citep{Pan2010-bv, Zamir2018-hr, Van_de_Ven2018-sd, Chen2018-uv, De_Lange2019-ib}.  
However, a consequence of the  generalized learning task defined here, any aspect of the setting for each task may change, each of these taxonomies is a low-dimensional projection of a high-dimensional landscape of potential continual learning problems.  We therefore propose a unifying  hierarchy (Figure~\ref{fig:schematic}), ranging from the simplest in-distribution learning tasks to the most complex biological learning tasks.  Note that even the lowest levels of the hierarchy contains many disparate kinds of tasks. For example, in-distribution learning contains as special cases essentially all of classical machine learning, including point estimation, hypothesis testing, supervised learning, unsupervised learning, casual learning, federated learning, streaming learning, online learning, and reinforcement learning.  Even with this list of different in-distribution learning problems, one can partition the space of problems in multiple ways: batch vs. online, supervised vs. unsupervised vs. reinforcement learning, perceptual vs. action, independent data samples or not, etc.  

Once we are faced with multiple data sets, as in transfer learning, already the landscape gets incredibly more complex. For example, given a pair of data sets, how are the query spaces related? Are they  the same space, overlapping spaces, one subspace is a strict subset of the other, or non-overlapping spaces?  The same question can be applied to the action space.  The statistical model (the set of admissible distributions) has similar questions, for example, if one data set sampled from a mixture distribution, where one of the components corresponds to the distribution of the other data set, are the distributions related by a rigid, linear, affine, or nonlinear transformation? 
Adding multiple tasks, as in multitask and meta-learning further complicates things.  For example, is it clear to the learner and/or the hypothesis which task each query is associated with? how much information about each tasks' setting is provided for any given query? Which components of the settings associated with each task differ from one another: (1) query space, (2) action space, (3) hypothesis space, (4) risk, (5) distribution, (6) statistical model, (7) evaluation distribution, (8) learner space, or (9) error?  
Adding dynamics, as in continual and lifelong learning further exacerbates these issues. For example, do the computational space and/or time constraints change for the learner and/or hypothesis? If so, in which ways?  Similarly, for some tasks data could arrive in batches, in others it could arrive sequentially, and the same is true for tasks.

The consequence of this inherent flexibility in defining generalized learning tasks complicates the literature.  Any given paper on `continual learning' could be solving one of many different kinds of problems.  Assume for the moment that a given paper is addressing a set of supervised learning classification tasks.  Given the nine different components above, and assuming only two different choices for each component, yields $2^9 = 512$ total possible continual learning classification problems, and different approaches will be designed typically to only address a very small subset of them. 
We therefore recommend greater specificity in manuscripts to clarify precisely the scope of the proposed learner and/or theory.  

Another consequence of this formalism is that it exposes that many previously proposed continual and lifelong  learning algorithms are, in fact, not respecting the computational complexity constraints to render them \textit{bona fide} continual learning.  Even for those that are, comparing algorithms with different computational complexity bounds is a bit like comparing apples to oranges. So, we advocate for more explicit theoretical and empirical investigations of computational complexity of these algorithms to understand their relative trade-offs.

\subsection{Limitations of the framework}

Our proposed framework attempts to unify in-distribution and out-of-distribution learning definitions; though it has several  limitations.  Perhaps most importantly from a machine learning perspective, we have not provided any theorems stating \textit{when} a given learner can solve a particular out-of-distribution learning task.  While  of the utmost importance to establish the theoretical  utility of this framework, we leave it to future work.  One of our motivating goals was to establish a learning framework that was sufficiently general to characterize both biological and machine learning.  However, this framework is inadequate for characterizing biological learning a few reasons. 

First, in biology, there are generally no explicit tasks.  Other forms of lifelong learning (e.g., \citet{Sutton2011Horde}) operate in a task-free setting, viewing the concept of a ``task'' as a convenient yet unrealistic construct. For example, what task are you doing right now?  Are you reading this paper, sitting balanced in your chair, listening to your surroundings, classifying typeset characters into letters and then words, recongizing speling errors, or some combination thereof? What precisely constitutes a task? \citet{Sutton2011Horde} operates by continually learning a set of functions (specifically, generalized value functions) that predict different aspects of the world and can be combined together hierarchically to achieve objectives. These functions are learned from a continual sensorimotor stream, without any extrinsic notion of tasks. To map such a task-free lifelong learner into our framework, we could view each of these prediction functions as a different intrinsic task that the agent must learn.  This permits the machinery we developed to equally apply to this scenario, with the important caveat that the tasks learned by the agent are not specified externally or \textit{a priori}.

Second, there are no discrete `samples' in biology; rather, biological agents are hit with a lifelong onslaught of data streams without a synchronizing clock.  At any given time a biological agent may be acquiring data, acting, and learning, or nothing at all.  
Third, for the most part, in biology, there is not simply one kind of unidimensional error. 
Instead, different kinds of inputs provide different kinds of affordances, such as oxygen, calories, and sleep~\citep{Gibson2014-or}.
In future work, we hope  to further bridge the gap between lifelong and biological learning by addressing these three limitations.

\subsection{The quest for artificial general intelligence}

It is unclear how problematic the differences are between our notion of lifelong learning and biological learning.  Many have recently argued that  modern machine learning and artificial intelligence is hitting a wall~\citep{Knight2019-zb}, or entering a new  winter~\citep{Piekniewski2018-tr}.  Various experts in machine learning and computer science  have proposed that the main bottlenecks to overcome include  causal learning~\citep{Pearl2019-bp}.  Other experts that lean more towards cognitive science have proposed  that the key bottleneck  is  symbolic reasoning~\citep{Marcus2020-vm} or the barrier of meaning~\citep{Mitchell2020-my}.  And still others believe that if we keep building larger and larger deep networks, with more data and bigger computers (and a few unspecified conceptual breakthroughs) we will eventually close the gap~\citep{Sutton2019-sr,Welling2019-if, Hao2020-xz}. Which of these beliefs will win in the end, if any, remains to be seen, though the fight continues on~\citep{debate}. We expect that  there will be many small victories from a wide diversity of individuals and approaches, that together will enable us to bridge the gap, including but not limited to bigger computers and data sets. 

\subsection{Concluding thoughts}

We were motivated to write this manuscript primarily for two reasons.  First, as we read the literature, we often found papers with proposed solutions to problems, but it was often unclear to us what problem was actually being solved, and how those problems are related to one another.  By virtue of standardizing an approach to quantifying out-of-distribution learning---via learning efficiency---we hope to be able to better understand the current state-of-the-art, and also advance beyond it.  Second, one particular way in which we hope to advance beyond the current state-of-the-art is by eclipsing biological learning in additional domains~\citep{Fjelland2020-xb}. We believe that a step towards realizing this dream includes establishing a formalism that is sufficiently flexible to be able to coherently evaluate learners in many different learning paradigms, including both biological and machine learners.  Indeed, arguably the crux of the gap between artificial general intelligence and natural intelligence is that natural biological learners are able to perform out-of-distribution learning well both within and across lifetimes.  We hope our proposed formalism can help clarify both where we are, and also where to go from here. 

\subsection*{Acknowledgements}
The authors thank the support of the National Science Foundation-Simons Research Collaborations on the Mathematical and Scientific Foundations of Deep Learning (MoDL, NSF grant 2031985). This work is graciously supported by the Defense Advanced Research Projects Agency (DARPA) Lifelong Learning Machines program through contracts FA8650-18-2-7834, HR0011-18-2-0025, and FA8750-18-C-0103. Research was partially supported by funding from Microsoft Research and the Kavli Neuroscience Discovery Institute. The authors are grateful for critical feedback from Yoshua Bengio,  Anirudh Goyal, Pratik Chaudhari, Rahul Ramesh, Raman Arora, Rene Vidal, Jeremias Sulam, Adam Charles, Anirudh Goyal, Timothy Verstynen, Konrad Kording, Jeffrey Dick,  Erik Peterson, Weiwei Yang,  Chris White, and Iris van Rooij. 

\bibliographystyle{unsrtnat}
\bibliography{biblifelonglearning.bib}



\end{document}